\documentclass{article}

\usepackage{times}
\usepackage{graphicx} 
\usepackage{subfigure} 

\usepackage{natbib}
 
\usepackage{algorithm}
\usepackage{algorithmic}

\usepackage{here}

\usepackage{hyperref}

\usepackage[accepted]{icml2015}

\icmltitlerunning{Dual Memory Architectures for Fast Deep Learning of Stream Data via an Online-Incremental-Transfer Strategy}

\begin{document} 

\twocolumn[
\icmltitle{Dual Memory Architectures for Fast Deep Learning of Stream Data\\
 via an Online-Incremental-Transfer Strategy}

\icmlauthor{Sang-Woo Lee}{slee@bi.snu.ac.kr}
\icmlauthor{Min-Oh Heo}{moheo@bi.snu.ac.kr}
\icmladdress{School of Computer Science and Engineering, Seoul National University}
\icmlauthor{Jiwon Kim}{g1.kim@navercorp.com}
\icmlauthor{Jeonghee Kim}{jeonghee.kim@navercorp.com}
\icmladdress{Naver Labs}
\icmlauthor{Byoung-Tak Zhang}{btzhang@bi.snu.ac.kr}
\icmladdress{School of Computer Science and Engineering, Seoul National University}


\vskip 0.3in
]

\begin{abstract}
The online learning of deep neural networks is an interesting problem of machine learning because, for example, major IT companies want to manage the information of the massive data uploaded on the web daily, and this technology can contribute to the next generation of lifelong learning.
We aim to train deep models from new data that consists of new classes, distributions, and tasks at minimal computational cost, which we call online deep learning. 
Unfortunately, deep neural network learning through classical online and incremental methods does not work well in both theory and practice.
In this paper, we introduce dual memory architectures for online incremental deep learning.
The proposed architecture consists of deep representation learners and fast learnable shallow kernel networks, both of which synergize to track the information of new data. 
During the training phase, we use various online, incremental ensemble, and transfer learning techniques in order to achieve lower error of the architecture.
On the MNIST, CIFAR-10, and ImageNet image recognition tasks, the proposed dual memory architectures performs much better than the classical online and incremental ensemble algorithm, and their accuracies are similar to that of the batch learner.
\end{abstract}

\section{Introduction}
Learning deep neural networks on new data from a potentially non-stationary stream is an interesting problem in the machine learning field for various reasons.
From the engineering perspective, major IT companies may want to update their services based on deep neural networks from the information of massive data uploaded to the web in real time.
From the artificial intelligence perspective, for example, we argue that online deep learning is the next probable step towards realizing the next generation of lifelong learning algorithms.
Lifelong learning is a problem of learning multiple consecutive tasks, and it is very important for creation of intelligent, general-purpose, and flexible machines \cite{thrun96,ruvolo13}. 
Online deep learning can have good properties from the perspective of lifelong learning because deep neural networks show good performance on recognition problems, and their transfer and multi-task learning problem \cite{heigold13,donahue14,yosinski14}.

However, it is difficult to train deep models in an online manner for several reasons. 
Most of all, the objective function of neural networks is not convex, thus online stochastic learning algorithms cannot guarantee convergence. 
Learning new data through neural networks often results in a loss of all previously acquired information, which is known as catastrophic forgetting. 
Because it is a disadvantageous constraint to learn one instance and then discard it in online learning, we can alleviate the constraint by memorizing a moderate amount of data (e.g., 10K).
We discover the online parameter of neural networks with an amount of data, which works reasonably for stationary data, but does not work well for non-stationary data.
On the other hand, if we have sufficient memory capacity, we can instead make an incremental ensemble of neural networks. 
Incremental ensemble learning refers to making a weak learner using new parts of an online dataset, and combining multiple weak learners to obtain better predictive performance.
There are several studies that use the incremental ensemble approach \cite{polikar01,oza01}.
In practice, however, a part of entire data is not sufficient for learning highly expressive representations of deep neural networks; therefore, the incremental ensemble approach alone does not work well, as illustrated in Section 3. 

To solve this problem, we use both online parametric and incremental structure learning. 
Because it is neither trivial nor easy to combine two approaches, we apply transfer learning to intermediate online and parameter learning.
This strategy, which we call an online-incremental-transfer strategy, is one of the key ideas for our proposed architecture.
For online incremental deep learning, we introduce the dual memory architecture that consists of the following two learning policies, and not simply a group of learning algorithms.
First, this architecture trains two memories -- one is an ensemble of deep neural networks, and the other are shallow kernel networks on deep neural networks. 
Two memories are designed for the different strategies. 
The ensemble of deep neural networks learns new information in order to adapt its representation, whereas the shallow kernel networks aim to manage non-stationary distribution and new classes in new data more rapidly.
Second, we use both online and incremental ensemble learning through the transfer learning technique. 
In particular, for example, we continually train a general model of the entire data seen in an online manner, and then, transfer to specific modules in order to incrementally generate an ensemble of neural networks. 
In our approach, online and incremental learning work together to achieve a lower error bound for the architecture.

The remainder of this paper is organized as follows. 
Section 2 briefly introduces the concept of the dual memory architecture. 
In Section 3 and 4, we propose and validate three specific examples of learning algorithms that satisfy the policies of the dual memory architecture. 
On the MNIST, CIFAR-10, and ImageNet image recognition tasks, the proposed algorithms performs much better than the classical online and incremental ensemble algorithm, and their accuracies are similar to that of the batch learner.
In Section 5, we summarize our arguments.

\section{Dual Memory Architectures}

\begin{figure}[ht]
\vskip 0.2in
\centering
\includegraphics[width=\columnwidth]{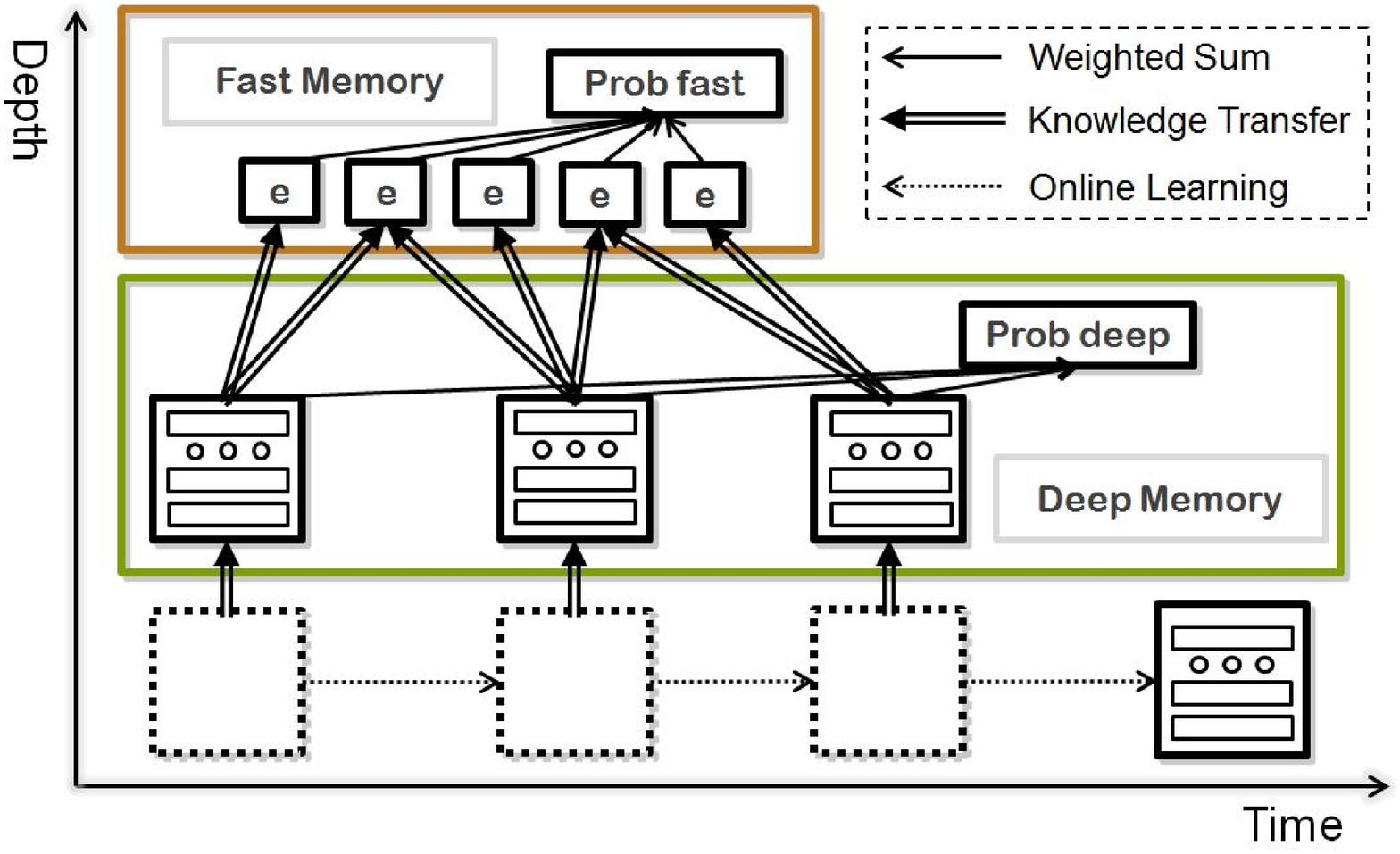}
\caption{An dual memory architecture.}
\end{figure}

In addition to the policies described in the previous section, we explain in general terms what dual memory architectures means, and discuss the type of algorithms that could be included in this framework.
However, this description is not restricted and can be extended beyond the given explanation in follow-up studies.
Dual memory architecture is the learnable system that consists of deep and fast memory, both of which are trained concurrently by using online, incremental, and transfer learning.

\begin{enumerate}
\item Dual memory architecture consists of an ensemble of neural networks and shallow kernel networks.
We call the former as ``deep memory,'' and the latter as ``fast memory'' (Figure 1).
\item Deep memory learns from new data in an online and incremental manner.
In deep memory learning, first, a general model is trained on the entire data it has seen in an online manner (first layer in Figure 1).
Second, the knowledge or parameter of the general model is transferred to incrementally generate an ensemble; weak neural network in the ensemble is specific for each data at a specific time (second layer in Figure 1) as clarified in Section 3.
\item Fast memory is on the deep memory. 
In other words, the inputs of the shallow kernel network are the hidden nodes of the higher layer of deep neural networks (third layer in Figure 1). 
The deep memory transfers its knowledge to the fast memory.
The fast memory learns from the new data in an online manner without much loss of accuracy compared with the batch learning process. 
However, batch learning, because of low computational cost in the parameter learning of shallow networks, can be used when higher accuracy is required.
\end{enumerate}

When new instances -- potentially a part of which has new distributions and additional classes -- arrive gradually, two memories ideally work as follows.
First, the weights of the fast memory are updated online with scant loss of the accuracy of the entire training data; for example, in the case of linear regression, no loss exists.
In this process, because of the transferability of the deep memory, the fast memory has remarkable performance, especially for new distributions and additional classes, as though the fast memory had already trained from many new instances with the same class and similar style \cite{donahue14}.
Second, representations of the deep memory also learn separately and more slowly from a stored moderate amount of data (e.g., 10K), especially because, when we need more data in order to make a new weak neural learner for an ensemble.
After a new weak neural learner is made, the fast memory makes new kernels that are functions of hidden values of both old and new weak learners.
In this procedure, the fast structure learning of the explicit kernel is particularly used in the paper.
As explained above, learning fast and slow is one of the mechanisms how the dual memory architectures work.

The other mechanism, online-incremental-transfer strategy, using both online stochastic and incremental learning through transfer learning technique, is explained in detail with examples.
In section 3, we discuss two specific algorithms for deep memory.
In section 4, we discuss one specific algorithm for fast memory.

\section{Online Incremental Learning Algorithms for Deep Memory}


For practical online learning from a massive amount of data, it is good to store a reasonable number of instances and discard those that appear less important for learning in the near future.
We refer to online learning as a parameter fine-tuning for new instances without retraining new model from an entire dataset that the model has seen ever.
As a type of practical online learning setting, we consider the ``mini-dataset-shift learning problem,'' which allows keeping at most $N_{subset}$ training examples in a storage for online learning (Algorithm 1).

\begin{algorithm}
\caption{Mini-Dataset-Shift Learning Problem}
\begin{algorithmic} 
\STATE{Initialize a model $\theta$ randomly.}
\REPEAT
\STATE{Get new data $D_{new}$.}
\STATE{Merge $D_{new}$ into the storage $D$  (i.e. $D \leftarrow D \bigcup D_{new}$).}
\STATE{Throw away some data in the storage to make $ |D| \leq N_{subset}$.}
\STATE{Train a model $\theta$ with $D$.}
\UNTIL forever
\end{algorithmic}
\end{algorithm}

To solve this problem, many researchers study incremental ensemble learning. 
We refer to incremental learning as structure learning for new instances; following the information of new data, a new structure is  made, and useless parts of the structure are removed.
Incremental ensemble learning, a type of both incremental and online learning, is referred to as combining multiple weak learners, each of which is trained on a part of that online dataset.
In this paper, our proposed algorithms are compared to the simple bagging algorithm or ``na{\"i}ve incremental ensemble.'' 
In this na{\"i}ve algorithm, for example, we train the first weak learner or neural network on the 1 -- 10,000$th$ data.
After that, the second neural network learns the 10,001 -- 20,000$th$ data.
Then, the third neural network learns the 20,001 -- 30,000$th$ data, and so on (if $N_{subset}$ is 10,000).
As mentioned later, however, this algorithm does not work well in our experiments.

\subsection{Mini-Batch-Shift Gradient Descent Ensemble}

First, we begin from an alternative approach -- online learning -- to complement the simple incremental ensemble approach.
The first step of our first algorithm involves using mini-batch gradient descent at each epoch with recent $N_{subset}$ training examples for accommodating $N_{new}$ new data.
We refer to this procedure as ``mini-batch-shift gradient descent.''
In this algorithm, for example, we first train on the 1 -- 10,000$th$ data with mini-batch gradient descent with sufficient epochs. 
After that, the model learns the 501 -- 10,500$th$ instances with one epoch. 
Then, the model learns the 1,001 -- 11,000$th$ instances with one epoch, and so on (if $N_{subset}$ is 10,000 and $N_{new}$ is 500). 

\begin{algorithm}
\caption{Mini-Batch-Shift Gradient Descent Ensemble}
\begin{algorithmic} 
\STATE{Collect first $N_{subset}$ new data $D_{first}$.}
\STATE{Learn a neural network $C$ with $D_{first}$ with enough epochs.}
\STATE{Put $D_{first}$ in the storage $D$ (i.e. $D \leftarrow D_{first}$).}
\REPEAT
\STATE{Collect $N_{new}$ new data $D_{new}$ such that $N_{new} < N_{subset}$.}
\STATE{Throw away the oldest $N_{new}$  instances in $D$.}
\STATE{Merge $D_{new}$ into $D$ (i.e. $D \leftarrow D \bigcup D_{new}$).}
\STATE{Train a general neural network $C$ with $D$ with one epoch.}
\IF{D is disjoint to the data used in $W_{prev}$}
\STATE{Initialize a new weak neural network $W_{new}$ by parameters of $C$.}
\STATE{Train $W_{new}$ with $D$ until converge.}
\STATE{Combine $W_{new}$ to a model $\theta$ (i.e. $\theta \leftarrow \theta \bigcup \{W_{new}\}$).}
\STATE{Refer to $W_{new}$ as $W_{prev}$ (i.e. $W_{prev} \leftarrow W_{new}$).}
\ENDIF
\UNTIL {forever}
\end{algorithmic}
\end{algorithm}

In Section 3.3, we show that mini-batch-shift gradient descent works well and outperforms the na{\"i}ve incremental ensemble.
Encouraged by this result, we apply mini-batch-shift gradient descent to incremental ensemble learning.
To combine online and incremental learning properly, we use the transfer learning technique.
Similar to the na{\"i}ve incremental ensemble, we train each neural network on each part of the online dataset.
Unlike the na{\"i}ve incremental ensemble, we transfer to each neural network from one trained on the entire data seen in an online manner.
We refer to the neural network trained in an online manner for the entire data as the general neural network $C$, whereas each weak neural network trained in a batch manner for each part of the online dataset is a weak neural network $W$.

To transfer from a general neural network $C$ to each weak neural network $W$, we use the \textit{initialize and fine-tune} approach suggested in \cite{yosinski14}.
The method we use is as follows: 1) initialize a target neural network with all parameters without the last softmax layer of a source neural network 2) fine-tune the entire target neural network.
Using this method, \cite{yosinski14} achieved 2.1\% improvement for transfer learning from one 500-classes to another 500-classes image classification task on the ImageNet dataset.
In the mini-batch-shift gradient descent ensemble, a general neural network $C$ trained by mini-batch-shift gradient descent is transferred to each weak neural network $W$ (Algorithm 2) and the ensemble of each weak learner $W$ is used for inference.
In mini-batch-shift gradient descent, we use one general neural network $C$ for inference, and do not make other neural networks.

\subsection{Neural Prior Ensemble}
Dual memory architecture is not just a specific learning procedure, but a framework for learning data streams. 
We introduce ``neural prior ensemble,'' another learning algorithm for deep memory.
In neural prior ensemble, a lastly trained weak neural network $W_{prev}$ takes the role of the general neural network $C$ used in the mini-batch-shift gradient descent, and it is transferred to a new weak neural network $W_{new}$ (Algorithm 3).
We refers to ``neural prior'' as the strategy for using the last neural network $W_{new}$ for inference, and neglect the previous neural networks in the next experiments section.

\begin{algorithm}
\caption{Neural Prior Ensemble}
\begin{algorithmic} 
\REPEAT
\STATE{Collect $N_{subset}$ new data $D_{new}$.}
\STATE{Initialize a new neural network $W_{new}$ by parameters of $W_{prev}$.}
\STATE{Train $W_{new}$ with $D_{new}$.}
\STATE{Combine a weak learner $W_{new}$ to a model. $\theta$ (i.e. $\theta \leftarrow \theta \bigcup \{W_{new}\}$)}
\STATE{Refer to $W_{new}$ as $W_{prev}$. (i.e. $W_{prev} \leftarrow W_{new}$)}
\UNTIL {forever}
\end{algorithmic}
\end{algorithm}

\begin{figure}[ht]
\vskip 0.2in
\centering
\includegraphics[width=\columnwidth]{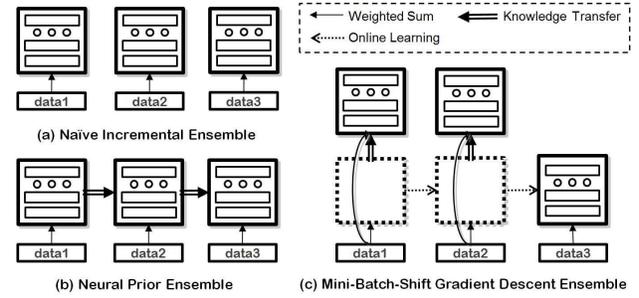}
\caption{Ensemble algorithms in the paper.}
\end{figure}

Figure 2 illustrates and summarizes ensemble algorithms for deep memory.
There is no knowledge transfer in na{\"i}ve incremental learning.
In mini-batch-shift gradient descent ensemble, a general neural network $C$ transfers their knowledge (first layer in Figure 2 (c)) to each weak neural network $W$ (second layer in Figure 2 (c)).
In neural prior ensemble, a lastly trained weak neural network $W_{prev}$ transfers their knowledge to a newly constructed neural network $W_{new}$.

\subsection{Experiments}

We evaluate the performance of the proposed algorithm on the MNIST, CIFAR-10, and ImageNet image object classification dataset. 
MNIST consists of 60,000 training and 10,000 test images, from 10 digit classes.
CIFAR-10 consists of 50,000 training and 10,000 test images, from 10 different object classes. 
ImageNet contains 1,281,167 labeled training images and 50,000 test images, with each image labeled with one of the 1,000 classes.
In experiments on ImageNet, however, we only use 500,000 images, which will be increased in future studies.
Thus, our experiments on ImageNet in the paper is somewhat disadvantageous because online incremental learning algorithms do worse if data is scarce in general.
We run various size of deep convolutional neural networks for each dataset using the demo code in MatConvNet, which is a MATLAB toolbox of convolutional neural networks \cite{vedaldi14}.
In our experiments, we do not aim to optimize performance, but rather to study online learnability on a standard architecture.

In the running of the mini-batch-shift gradient descent, we set the learning rate proportional to $1/\sqrt{t}$, where $t$ is a variable proportional to the number of entire data that the model has ever seen.
In the other training algorithms, including the batch learning and the neural prior, we first set the learning rate $10^{-2}$ and drop it by a constant factor -- in our experiments, 10 -- at some predifined steps.
In entire experiments, we exploit the momentum of the fast training of neural networks; without momentum, we could not reach the reasonable local minima within a moderate amount of epochs in our experiments.

\begin{figure}[H]
\centerline{\includegraphics[width=\columnwidth]{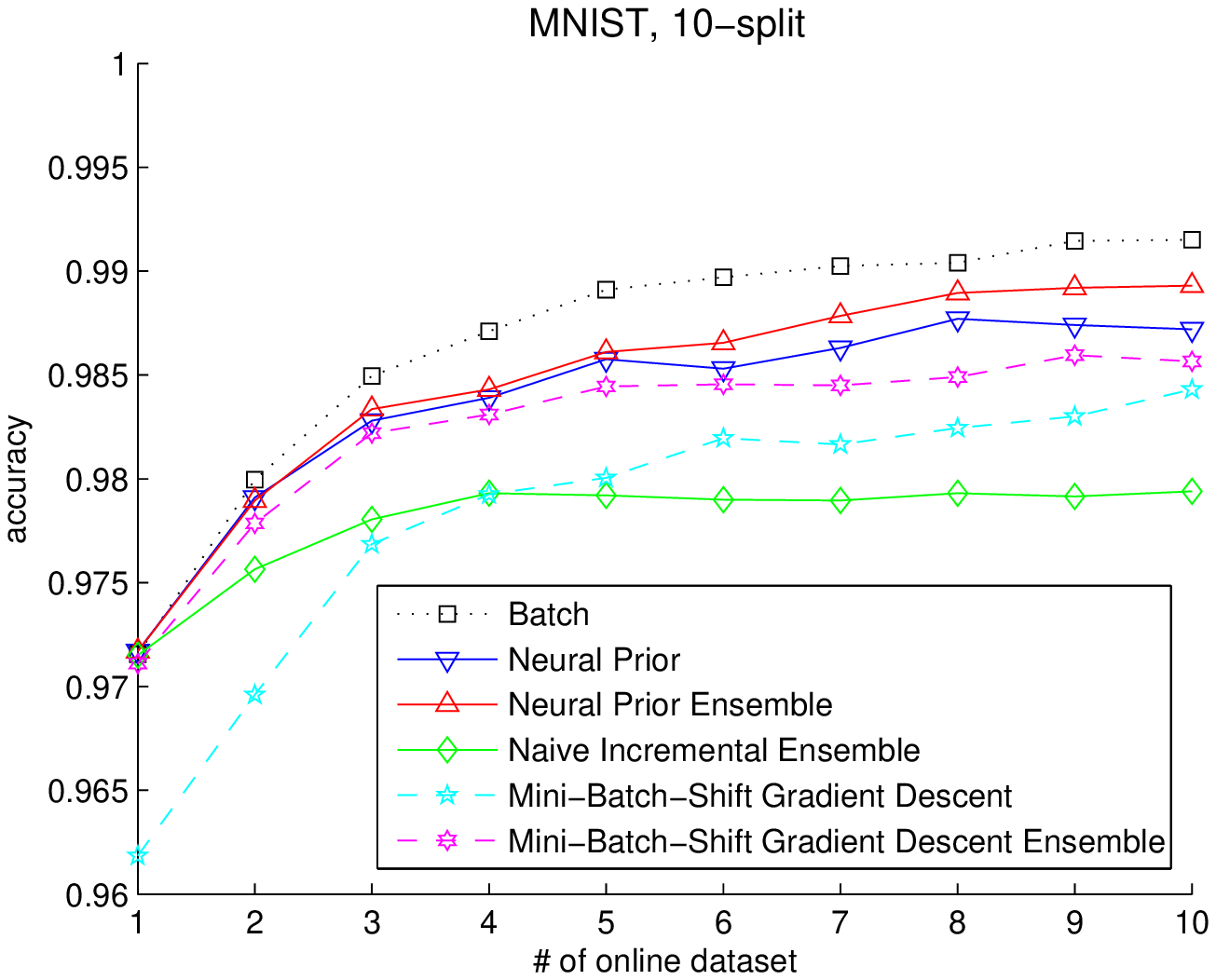}}
\centerline{\includegraphics[width=\columnwidth]{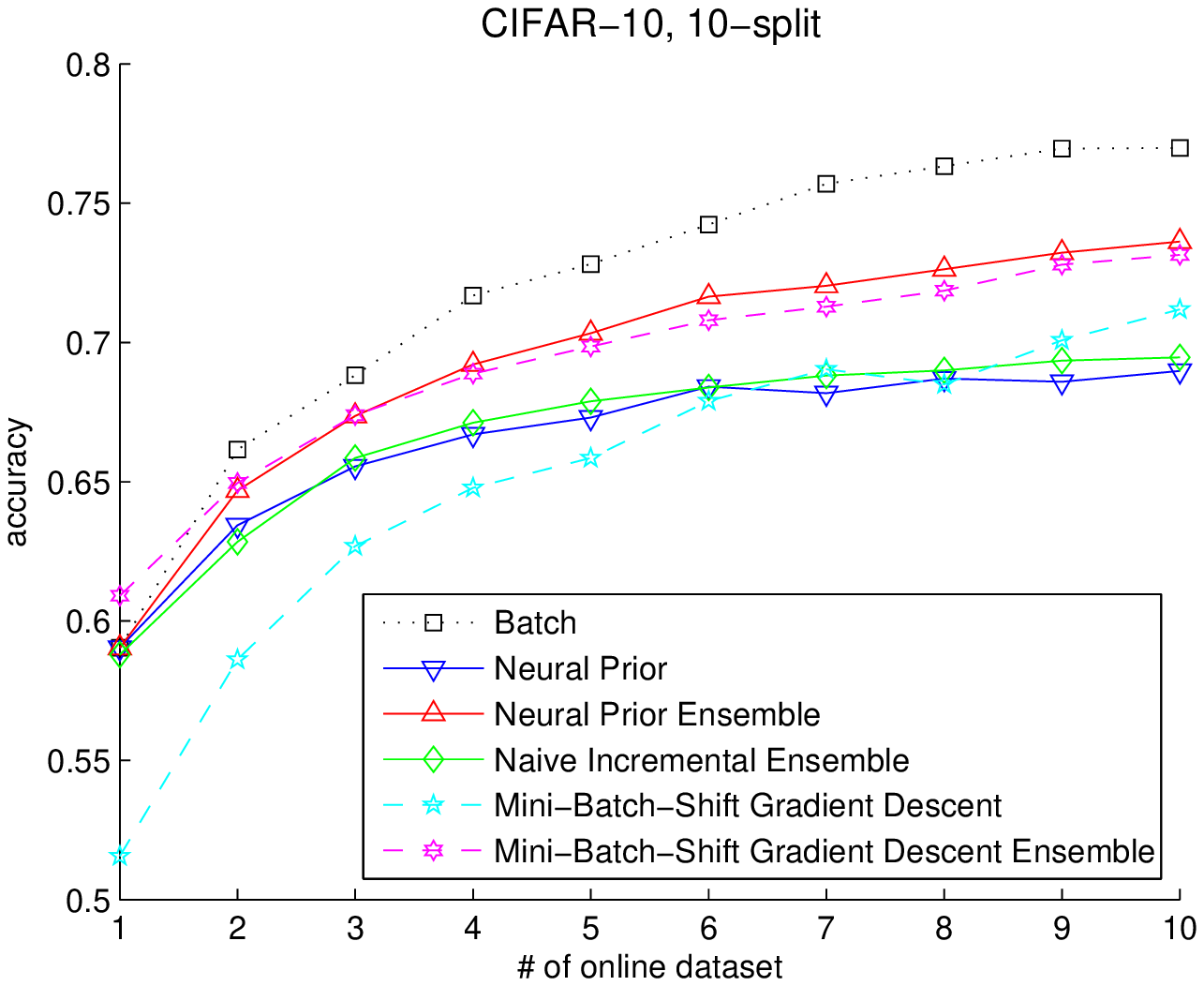}}
\centerline{\includegraphics[width=\columnwidth]{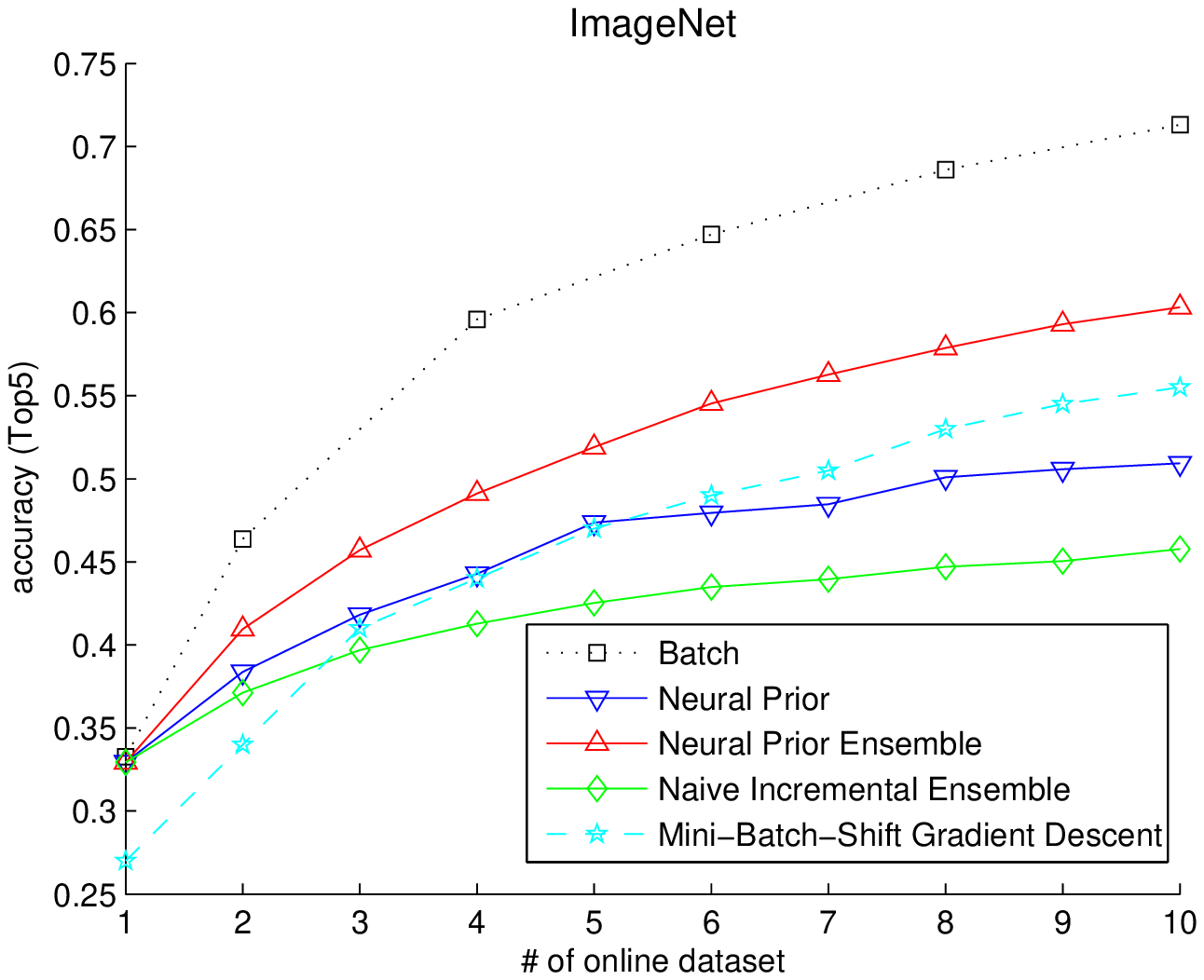}}
\caption{Results of 10-split experiments on MNIST, CIFAR-10, and ImageNet.}
\end{figure}\

The main results on deep memory models are shown in Figure 3.
We randomly split the entire training data into the 10 online dataset to make the distribution of the data stream stationary; we call this setting `10-split experiments'. 
In this setting, we maintain $1/10$ of each entire dataset as the number of training examples $N_{memory}$ in the storage.

First, these results show that mini-dataset-shift learning algorithms with a single general neural network -- i.e. the mini-batch-shift gradient descent and the neural prior -- outperform the na{\"i}ve incremental ensemble.
In other words, the online learning of a neural network with an amount ($N_{memory}$) of stored data is better than simply bagging each weak neural network with the same amount of data.
Our experiments show that learning a part of the entire data is not sufficient to make highly expressive representations of deep neural networks.

Meanwhile, the lower accuracies in the early phase of the mini-batch-shift gradient descent are conspicuous in each figure because we remain as a relatively high learning rate that prevents efficient fine-tuning.
We improved the performance of the early phase with batch-style learning of the first online dataset without loss of the accuracy of the latter phase in other experiments not shown in the figures.
The figure also illustrates that ensemble algorithms for deep memory -- i.e. mini-batch-shift gradient descent ensemble and neural prior ensemble -- perform better than algorithms with a single neural network.
Regardless of the improvement, it is a burden to increase the memory and inference time proportional to data size in the ensemble approach.

\begin{figure}[ht]
\centering
	\includegraphics[width=\columnwidth]{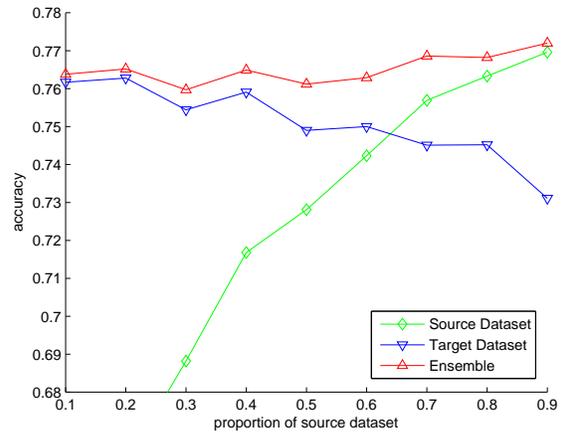}
\caption{Results of two-split experiments on CIFAR-10}
\end{figure}

When the data distribution is stationary, however, we found that maintaining a small number of neural networks does not decrease accuracy significantly.
In our experiment, for example, selecting three over ten neural networks at the end of learning in the neural prior ensemble simply decreases the absolute error to less than 1\%.

The performances of the proposed online learner may seem insufficient compared with the batch learner.
However, by alleviating the condition, the entire dataset is divided into two online datasets, the performance losses of the proposed ensemble decrease.
Figure 4 show the results on CIFAR-10 split into two online datasets with various proportions of the source and target parts.

\section{Online Incremental Learning Algorithms for Fast Memory}
\subsection{Shallow Kernel Networks on the Neural Networks}
We introduce the fast memory; shallow kernel networks on the neural networks.
In dual memory architectures, the input features of shallow kernel networks we used as fast memory are the activation of deep neural networks.
Complementing the dual memory, the fast memory plays two important roles for treating stream data.
First, a fast memory integrates the information distributed in each neural networks of ensemble.
On the non-stationary data stream, not only proposed mini-dataset-shift learning algorithm of a single neural network but also ensemble learning algorithm for deep memory does not work well.
Training fast memory with entire training data makes much better performance than deep memory alone, in particular, when new data includes new distributions and additional classes.
It is quite practical, because of low computational costs on parameter learning of shallow networks.
Second, fast memory can be updated from each one new instance, with a small amount of calculation until the features remain unchanged.
It does not require without much gain of loss function comparing to the batch counterpart; in case of the linear regression, loseless.
Learning deep memory needs expensive computational costs on inference and backpropagation in deep neural networks, even if deep memory is trained through the online learning algorithm we proposed.

\subsection{Multiplicative Hypernetworks}
In this section, we introduce a multiplicative hypernetwork (mHN) as an example of fast memory.
This model is inspired by the sparse population coding model \cite{zhang12} and it is revised to be fit to the classification task we want to solve.
We choose mHNs for their good online learnability via sparse well-shared kernels among classes.
However, there are alternative choices, e.g., a support vector machine (SVM) \cite{liu08}, and an efficient lifelong learning algorithm (ELLA) \cite{zhou12}, among which SVM is our comparative model.
mHNs are shallow kernel networks that use a multiplicative function as a explicit kernel $\phi = \{\phi^{(1)}, ..., \phi^{(P)}\}^T$ where 
\begin{center}
$\phi^{(p)}(v,y) = (v_{(p,1)} \times ... \times v_{(p,K_p)})$ $\&$ $\delta(y) $.
\end{center}
$\times$ denotes the scalar multiplication and $\delta$ denotes the indicator function. 
$v$ is the input feature of mHNs, which is also the activation of deep neural networks, and $y$ is the target class.
$\{v_{(p,1)}, ..., v_{(p,K_p)}\}$ is the set of variables used in $pth$ kernel.
$K_p$ is the order, or the number of variable used in $pth$ kernel; in this paper $K_p =$ 2.
In the training of parameters that correspond to kernels, we obtain weights by least-mean-square or linear regression formulation. 
We use one-vs.-rest strategy for classification; i.e., the number of linear regressions is the same as that of the class, and the score of each linear regression model is evaluated.
This setting guarantees loseless weight update until the features remain unchanged.

\begin{center} 
$P_0 = I, B_0 = 0$\\
$P_t = P_{t-1}[I-\frac{\phi_t\phi_t^TP_{t-1}}{1+\phi_t^TP_{t-1}\phi_t}]$\\
$B_t = B_{t-1} + \phi^T_ty_t$\\
$w^*_t = P_tB_t$
\end{center}

Where $y_t$ is the Boolean scalar whether the class is true or false (i.e., 0 or 1), and $\phi_t$ is a kernel vector of $tth$ instance, the form of kernel $\phi$ can have various features, and the search space of the set of kernels is an exponential of an exponential.
To tackle this problem, we use evolutionary approach to find a near optimal set of kernels. 
We randomly make new kernels and discard some kernels less relevant.
Algorithm 4 explains the online learning procedure of  multiplicative hypernetworks.

\begin{algorithm}
\caption{Learning Multiplicative Hypernetworks}
\begin{algorithmic} 
\REPEAT
\STATE{Get a new instance $d_{new}$.}
\IF{$d_{new}$ includes new raw feature}
\STATE{Make new kernels $\phi_{new}$ including the values of new feature explicitly.}
\STATE{Merge $\phi_{new}$ into kernels of model $\phi$.}
\STATE{Fine-tune weights of kernels $W$ of $\phi$ with the storage $D$.} 
\STATE{Discard some kernels in $\phi$ which seem to be less relevant to target value.}
\ENDIF
\STATE{Update $W$ with $d_{new}$.} 
\STATE{Combine $d_{new}$ to $D$  (i.e. $D \leftarrow D \bigcup \{d_{new}\}$).}
\STATE{Throw away some data in the storage seem to be less important for learning in the near future.}
\UNTIL {forever}
\end{algorithmic}
\end{algorithm}

\subsection{Experiments}
We evaluate the performance of the proposed fast memory learning algorithm with convolutional neural networks (CNNs) and mHNs on CIFAR-10 dataset.
In this setting, we split the entire training data into the 10 online datasets with non-stationary distribution of the class.
In particular, the first online dataset consists of 40\% of class 1, 40\% of class 2, and 20\% of class 3 data.
The second online dataset consists of 40\% of class 1, and 20\% of class 2 -- 5 data.
The third online dataset consists of 20\% each of class 1 -- 5 data.
The fourth online dataset consists of 20\% each of class 2 -- 6 data, and so on.
We maintain $1/10$ of entire dataset as the number of training examples $N_{memory}$ in the storage.
We mainly validate mHNs on the deep neural networks where the neural prior ensemble is used for learning deep memory.
We train mHNs in strictly online manner until new weak learner of ensemble is added; otherwise we allow the model to use previous data it has ever seen. 
It is limitation of our works and will be discussed and improved in follow-up studies.
 
\begin{figure}[t]
\centering
\includegraphics[width=\columnwidth]{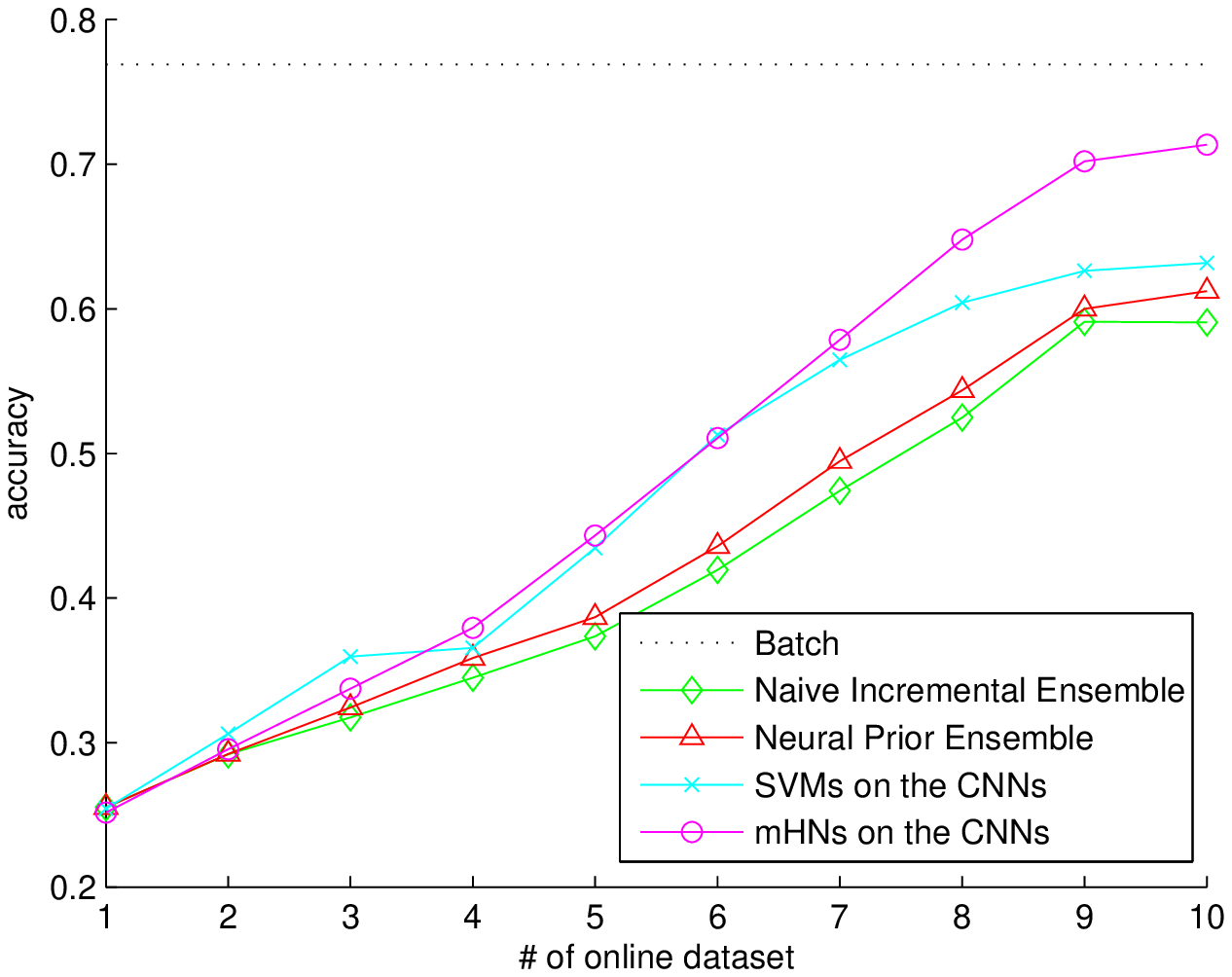}
\includegraphics[width=\columnwidth]{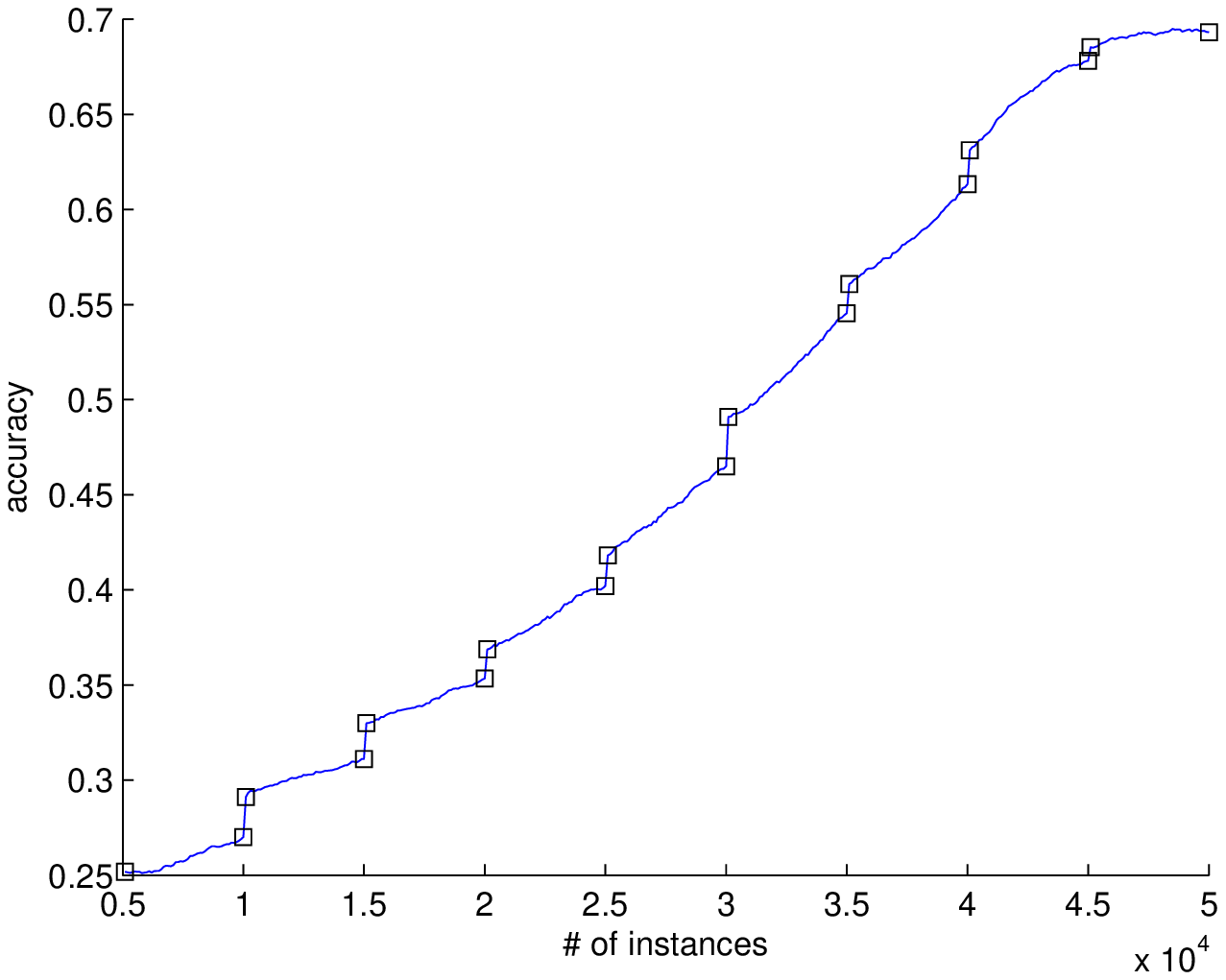}
\caption{Experimental results on CIFAR-10. (Top) the accuracy of various learning algorithms on non-stationary data. (Bottom) the accuracy of the mHN on the CNNs plotted at the every time the one new instance comes.}
\end{figure}

The main experiment results on the fast memory models are shown in Figure 5.
We use neural prior ensemble for deep memory when we validate the fast memory algorithms.
Although not illustrated in the figure, the mini-batch-shift gradient descent and neural prior converge rapidly with the new online dataset and forget the information of old online datasets, as indicated by the research on catastrophic forgetting.
Thus, the performance of the deep memory algorithm on a single neural network does not exceed 50\% because each online dataset does not include more than 50\% of the classes.
The accuracy of the neural prior ensemble exceeds 60\%, but it is not sufficient compared with that of the batch learner.
The fast memory algorithms -- the mHNs on the CNNs, the SVMs on the CNNs -- work better than a single deep memory algorithm.
A difference of the performance between mHNs and SVMs in the latter phase is conspicuous in the figure, whose meaning and generality is discussed in follow-up studies.


The bottom subfigure of Figure 5 shows the performance of the mHNs on the CNNs plotted at the exact time that one new instance arrives. 
Small squares note the points that before and after a new weak neural network is made by the neural prior ensemble algorithm.
The figure shows not only that fast memory rapidly learns from each instance of the data stream, but also that the learning of the weak deep neural networks is also required.
In our experiments, learning mHNs is approximately 100 times faster than learning weak neural networks on average.

\section{Conclusion}


We introduced dual memory architectures to train deep representative systems without much loss of online learnability.
In this paper, we studied some properties of online deep learning.
First, deep neural networks have online learnability on large-scale object classification tasks for stationary data stream.
Second, for extreme non-stationary data stream, deep neural networks forget what they learned previously; therefore, making a new module incrementally can alleviate this problem. 
Third, by transferring knowledge from an old module to a new module, the performance of online learning systems is increased.
Fourth, by placing shallow kernel networks on deep neural networks, the online learnability of the architecture is enhanced.

In this paper, numerous practical and theoretical issues are revealed, which will be soon discovered in our follow-up studies.
We hope these issues will be discussed in the workshop.

\section*{Acknowledgments} 
This work was supported by the Naver Labs. 
This work was partly supported by the NRF grant funded by the Korea government (MSIP) (NRF-2010-0017734-Videome) and the IITP grant funded by the Korea government (MSIP) (R0126-15-1072-SW.StarLab, 10035348-mLife, 10044009-HRI.MESSI).


\bibliography{example_paper}
\bibliographystyle{icml2015}

\end{document}